\pdfoutput=1

\documentclass[11pt]{article}
\usepackage{multirow}
\usepackage{colortbl}
\usepackage[table,xcdraw]{xcolor}

\definecolor{lightyellow}{rgb}{1.0, 1.0, 0.6}
\usepackage{ulem}
\usepackage{tcolorbox}
\usepackage{xcolor}
\usepackage[final]{acl}
\usepackage{booktabs} 
\usepackage{listings}
\usepackage{times}
\usepackage{fontawesome}
\usepackage{latexsym}
\usepackage{makecell}
\usepackage[T1]{fontenc}

\usepackage[utf8]{inputenc}
\usepackage{color}

\definecolor{algblue}{RGB}{23, 100, 171}
\definecolor{comprblue}{RGB}{208, 225, 242}
\definecolor{mathblue}{RGB}{148, 196, 223}
\definecolor{logicblue}{RGB}{74, 152, 201}
\definecolor{darkgreen}{rgb}{0.0, 0.5, 0.0}
\definecolor{lightblue}{RGB}{173, 216, 230}
\definecolor{lightgrey}{RGB}{211, 211, 211}
\usepackage{microtype}

\usepackage{inconsolata}

\usepackage{graphicx}

\title{Assessing Dialect Fairness and Robustness of\\ Large Language Models in Reasoning Tasks}
\author{Fangru Lin$^{1}$\thanks{This work was conducted during an internship at Microsoft Research. Contact: fangru.lin@ling-phil.ox.ac.uk, dawnmsg941107@gmail.com} \quad
  Shaoguang Mao$^{2}$ \quad Emanuele La Malfa$^{1,3}$ \quad Valentin Hofmann$^{4, 5}$ \\
  \textbf{Adrian de Wynter$^{6,7}$ \quad Xun Wang$^{6}$ \quad Si-Qing Chen$^{6}$}\\
  \textbf{Michael Wooldridge$^{1,3}$ \quad Janet B. Pierrehumbert$^{1}$ \quad Furu Wei$^{2}$}\\
  \begin{tabular}{c}
    $^{1}$University of Oxford \quad
    $^{2}$Microsoft Research \quad
    $^{3}$The Alan Turing Institute\\
    $^{4}$Allen Institute for AI \quad
    $^{5}$University of Washington\\$^{6}$Microsoft Corporation \quad $^{7}$University of York
  \end{tabular}}

\begin{document}
\maketitle
\begin{abstract}
Language is not monolithic. While benchmarks, including those designed for multiple languages, are often used as proxies to evaluate the performance of Large Language Models (LLMs), they tend to overlook the nuances of within-language variation and thus fail to model the experience of speakers of non-standard dialects. 
Focusing on African American Vernacular English (AAVE), we present the first study aimed at objectively assessing the fairness and robustness of LLMs in handling dialects across canonical reasoning tasks, including algorithm, math, logic, and integrated reasoning. We introduce \textbf{ReDial} (\textbf{Re}asoning with \textbf{Dial}ect Queries), a benchmark containing 1.2K+ parallel query pairs in Standardized English and AAVE. We hire AAVE speakers, including experts with computer science backgrounds, to rewrite seven popular benchmarks,
such as HumanEval and GSM8K. With ReDial, we evaluate widely used LLMs, including GPT, Claude, Llama, Mistral, and the Phi model families. Our findings reveal that \textbf{almost all of these widely used models show significant brittleness and unfairness to queries in AAVE}. Our work establishes a systematic and objective framework for analyzing LLM bias in dialectal queries. Moreover, it highlights how mainstream LLMs provide unfair service to dialect speakers in reasoning tasks, laying a critical foundation for future research.\footnote{Code and data can be accessed \href{https://github.com/fangru-lin/redial_dialect_robustness_fairness}{here}.}
\end{abstract}

\section{Introduction}
\label{sec:introduction}
Over the last few decades, linguistic research has firmly established that language naturally varies along social, geographic, and demographic dimensions \citep{chambers1998dialectology}. Such \textit{dialectal} variation is one of the most salient forms of linguistic diversity. Speakers of “non-standard” dialects are
known to experience implicit and explicit discrimination in everyday situations, including housing, education, employment, and the criminal justice system \citep{baugh2005linguistic, adger2014dialects, rickford2016language, drozdzowicz2024complexities}. As Large Language Models (LLMs) increasingly serve a broad and rapidly expanding user base \citep{author2023chatgpt, la2024language}, it is critical to understand how they interact with diverse linguistic communities. 

In this work, we examine LLMs' \textbf{dialect robustness and fairness}. For \textbf{robustness}, adversarial robustness provides a consolidated framework to test LLMs on slight variations of existing tasks~\citep{DBLP:journals/corr/abs-2108-12237, jin-etal-2023-adversarial}. Dialects reformulate a problem while maintaining its semantics, i.e., they test what has been referred to as \textit{semantic robustness} \citep{Malfa_Kwiatkowska_2022}. For \textbf{fairness}, recent research has demonstrated that LLMs exhibit biases against non-standard dialect queries, predominantly assessed in language and social analysis tasks \citep{sap-etal-2019-risk, ziems-etal-2023-multi, hofmann2024dialect}. Equally relevant, yet less studied, are tasks that require reasoning abilities for problem-solving, decision-making, and critical thinking \citep{wason1972psychology,huth2004logic,huang2022towards, qiao2022reasoning}. For instance, algorithm-related tasks (e.g., generation, debugging, etc.) figure prominently in real user queries, as reflected by their first place on the ArenaHard quality board \citep{li2024crowdsourced} and their third place on the WildChat frequency board \citep{zhao2024wildchat}. 

\begin{figure*}[t]
    \centering
    \includegraphics[trim=10 40 15 40, clip, width=0.8\linewidth]{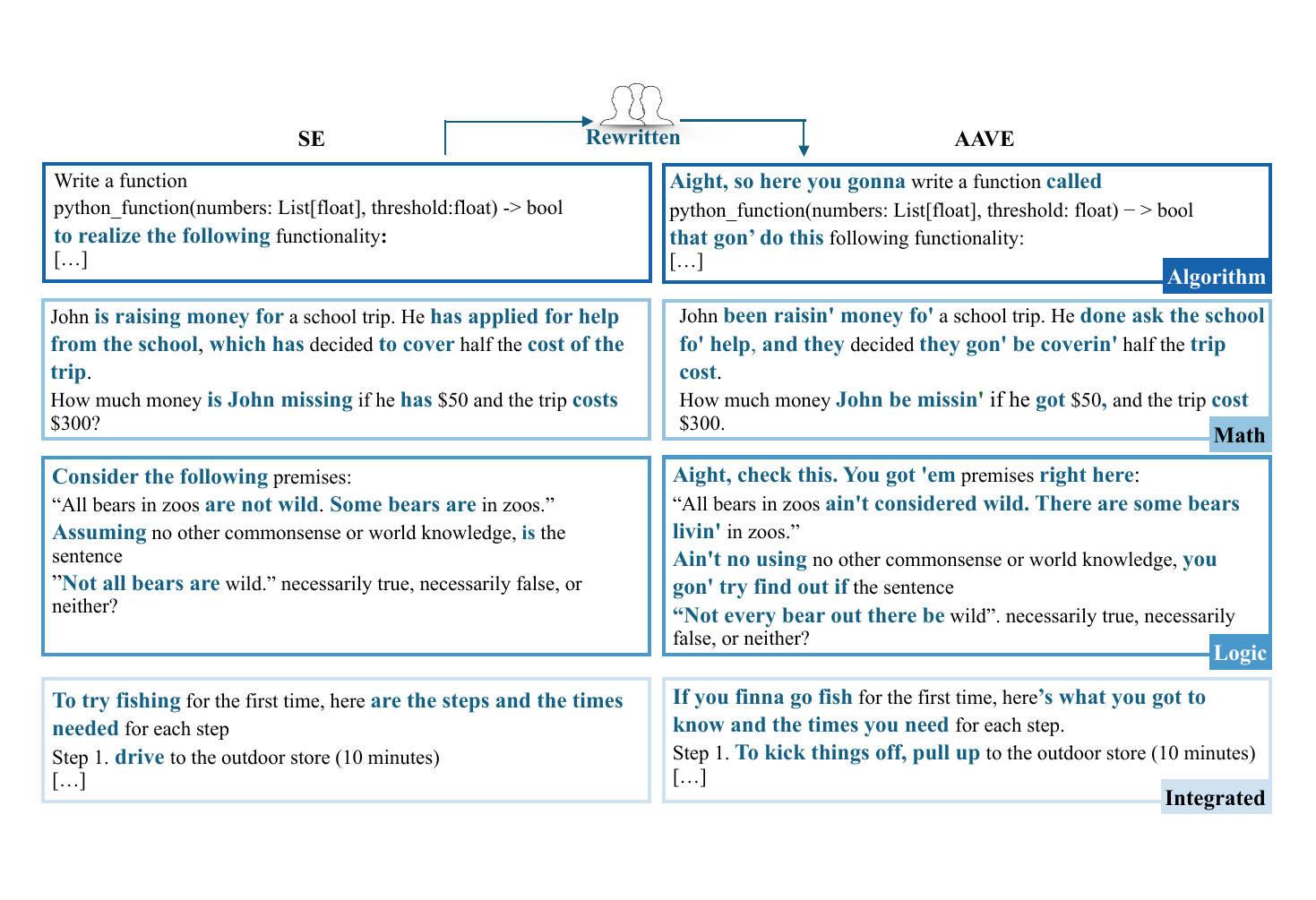}
    \caption{ReDial is a dialect reasoning benchmark composed of 1.2K+ SE-AAVE parallel queries. Its source data comes from existing benchmarks in SE. AAVE speakers are hired to rewrite each instance in their dialect but preserve their original intent, meaning, and ground truth output label to form their AAVE counterparts. Highlighted parts in blue are major differences in AAVE rewriting compared to SE.}
    \label{fig:redial_egs}
\end{figure*}

However, existing dialectal benchmarks \citep[e.g.,][]{ziems-etal-2023-multi} do not cover these tasks, and current popular reasoning benchmarks such as HumanEval \citep{chen2021evaluating} and GSM8K \citep{cobbe2021gsm8k} are constructed in Standardized English (SE). This could disadvantage dialect speakers in real-world applications like educational assessment \citep{gonzalez2021artificial}, personalized recommendation \citep{kantharuban2024stereotype}, and even multimodal tasks \citep[e.g., voice assistants;][]{10.1093/applin/amac066}, ultimately forcing them to shift their language styles to SE \citep{cunningham2024understanding} in order to access the full benefits of modern technologies, even though they prefer to use their own dialects \cite{blaschke-etal-2024-dialect}.

We present the first study that \textbf{systematically and objectively evaluates LLM fairness and robustness in reasoning tasks expressed in a non-standard dialect}. We choose AAVE since around $33$ million people worldwide and approximately $80\%$ of African Americans in the United States speak AAVE, with reports of discriminative behaviors in various scenarios~\citep{lippi1997we, purnell1999perceptual, massey2001use, grogger2011speech}. Moreover, relevant research shows that many speakers of English have difficulty understanding AAVE \cite{rickford2016language}. With the objective of making sure that users can use the language style they prefer instead of being constrained by the preference of the language model service, we consider that there is a need to separately consider AAVE from SE in evaluating LLMs. We hire AAVE speakers to manually rewrite instances from seven popular SE reasoning benchmarks into AAVE (Section~\ref{sec:data_source}). Our approach has unique advantages compared with prior works that either (i) rely on predefined lexical or morphosyntactic transformation rules \citep{ziems2022value, ziems-etal-2023-multi}, which may overlook subtle contextual nuances, or (ii) use LLMs as translators \citep{gupta2024aavenue}, which may have the very biases that our research wants to unveil \citep{fleisig2024linguistic, smith2024standard}.

\begin{table*}[t]
    \centering
    \resizebox{0.9\textwidth}{!}{
    \begin{tabular}{c|c|c|c|c|c|c|c|c}
    \toprule
       \textbf{Category} &\multicolumn{2}{|c}{\textbf{Algorithm} (25.7\%)}&\multicolumn{2}{|c}{\textbf{Logic} (29.8\%)}&\multicolumn{2}{|c|}{\textbf{Math} (24.7\%)}&\textbf{Integrated} (19.7\%)&\textbf{Total}\\
       \midrule
       \textbf{Source} &HumanEval&MBPP&LogicBench&Folio&GSM8K&SVAMP&AsyncHow& -\vspace{0.8mm} \\
       \textbf{Size} &164&150&200&162&150&150&240&1,216\\
       \bottomrule
    \end{tabular}%
}
    \caption{ReDial contains 1,216 fully-annotated parallel prompts for four categories, drawn from seven data sources. Each category’s contribution to the total is reported in percentage points in brackets.}
    \label{tab:ReDial_overview}
\end{table*}

We introduce \textbf{ReDial} (\textbf{Re}asoning with \textbf{Dial}ect Queries), \textbf{the first high-quality, end-to-end human-annotated SE-AAVE parallel dataset for reasoning tasks} (Section~\ref{sec:dataset}). ReDial contains over 1.2K SE-AAVE prompt pairs covering four canonical reasoning categories: \textit{algorithm}, \textit{math}, \textit{logic}, and \textit{integrated reasoning} (tasks that require composing multiple reasoning skills). By anchoring these queries to known correct answers and employing human-based rewriting, ReDial furnishes an objective measure of dialect fairness and robustness. It also avoids the pitfalls of LLM-based evaluations, which can be inherently biased \citep{zheng2023judging, chen2024humans, shi2024judging}.

Using ReDial, we benchmark widely used LLMs, including GPT-o1, GPT-4o, Claude-3.5-Sonnet, Llama-3.1-70B-Instruct, and others (Section~\ref{sec:experiment}). We find that almost all models experience statistically significant performance drops on AAVE prompts, despite their semantic equivalence to their SE counterparts. On average, we observe a relative performance reduction of more than 10\%. This discrepancy persists even with advanced prompting techniques like Chain of Thought (CoT) prompting \citep{kojima2022large, wei2022chain}, indicating that current LLMs are both brittle and unfair with dialectal inputs.

To understand these gaps, we further analyze potential causes. Our analysis reveals that the brittleness of LLMs with AAVE prompts arises from a combination of dialect-specific morphosyntactic features and nuanced conversational norms. Experiments with synthetic perturbations and AAVE-specific feature injections show that while these factors contribute to performance degradation, they fail to replicate the severity observed with human-annotated data. This highlights the limitations of rule-based transformations and the critical need for high-quality, context-rich datasets like ReDial to evaluate LLM fairness and robustness effectively.

\noindent In summary, in this paper:
    
\begin{enumerate}
    \item We introduce ReDial, the first high-quality, end-to-end human-annotated AAVE-SE parallel reasoning benchmark spanning four foundational reasoning tasks.
    \item We show that leading LLMs exhibit significant unfairness and brittleness on AAVE prompts compared to their SE counterparts.
    \item We identify that the brittleness of LLMs with AAVE prompts stems from a combination of dialect-specific morphosyntactic features and nuanced conversational norms, which cannot be captured by synthetic transformations.
\end{enumerate}

\section{Dataset}
\label{sec:dataset}
\textbf{ReDial} (\textbf{Re}asoning with \textbf{Dial}ect Queries) is a benchmark of more than 1.2K parallel Standard English–African American Vernacular English (SE-AAVE) query pairs. Table~\ref{tab:ReDial_overview} provides an overview of the distribution, and Figure~\ref{fig:redial_egs} along with Appendix~\ref{sec:redial_samples} present illustrative examples. 

Following \citet{zhu2023dyval}, ReDial includes canonical reasoning tasks—\textbf{algorithm}, \textbf{logic}, and \textbf{math}. We additionally consider \textbf{integrated reasoning} as a compositional task requiring multiple skills. The task formulation of ReDial is linguistically diverse, addresses cornerstone problems in human reasoning, and is of particular interest as it is challenging for LLMs. 

We first describe the data sources and sampling strategies (Section~\ref{sec:data_source}), and then detail the AAVE rewriting and validation processes that ensure high data quality (Section~\ref{sec:annotation_quality_check}).

\subsection{Data Sourcing}
\label{sec:data_source}

We construct a highly curated dataset by drawing upon seven established benchmarks covering different aspects of reasoning. We purposefully select the benchmarks that can capture the needs of real-world applications.

For each source, we provide key references, task descriptions, and sample sizes. Additional examples can be found in Appendix~\ref{sec:per_dataset_illustration_appendix}.

\begin{figure*}[t]
    \centering
    \includegraphics[trim=30 50 45 150, clip, width=0.8\linewidth]{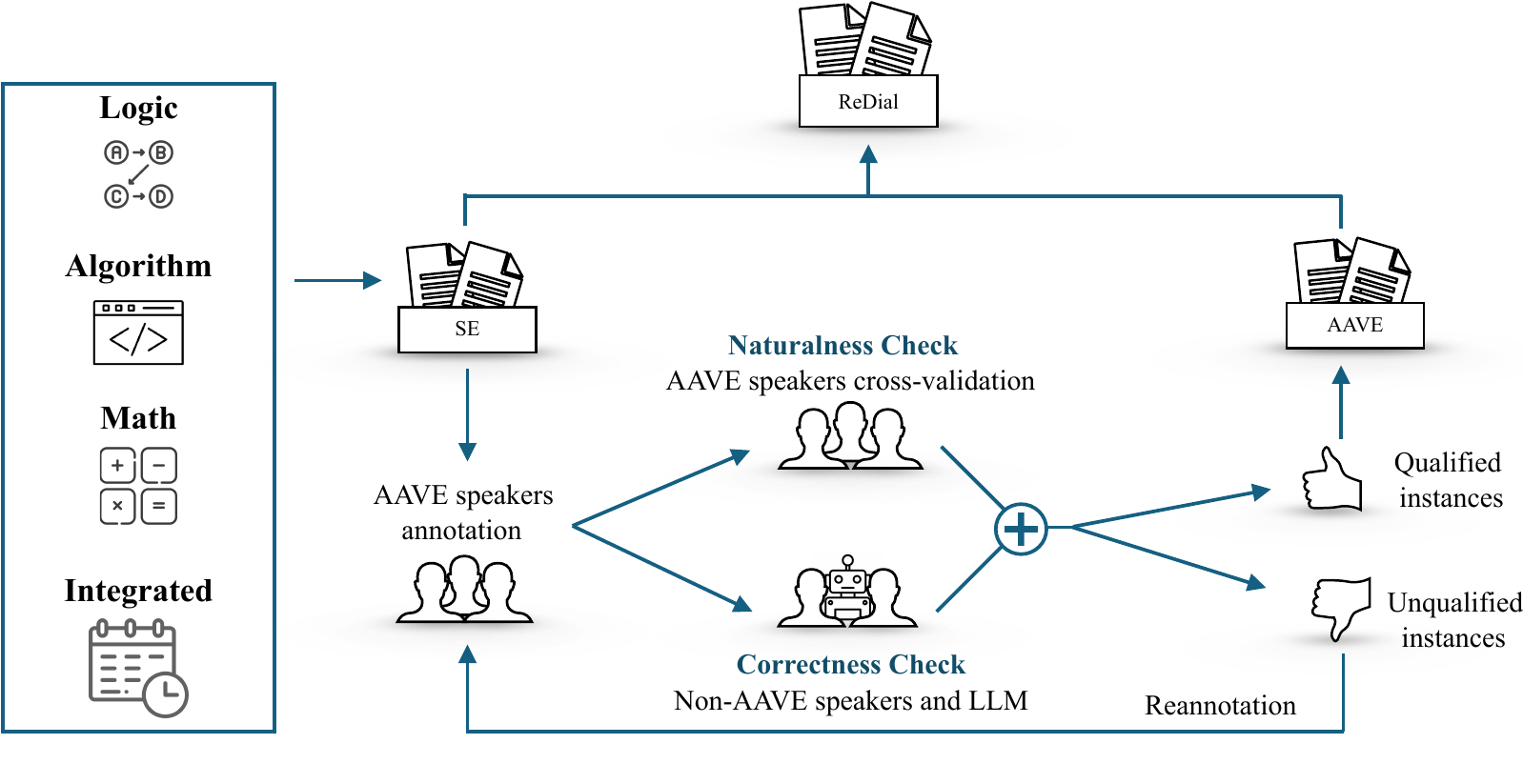}
    \caption{Annotation and cross-validation of ReDial instances. We first sample instances from datasets of four canonical reasoning tasks to compose the source data, then we hire AAVE speakers to rewrite the instances in their dialect. To ensure the quality, we conduct a \textbf{naturalness check} by AAVE speakers and a \textbf{correctness check} by non-AAVE speakers and LLM. We reannotate instances that do not pass the quality checks and iterate the process until the data meets our criteria. Finally, we combine the source data and AAVE rewriting to obtain ReDial.}
    \label{fig:annotation-redial}
\end{figure*}

\colorbox{algblue}{\textcolor{white}{Algorithm}} \textbf{HumanEval} \citep{chen2021evaluating} consists of 164 human-written code completion instances. We convert and include all these code completion headings into instruction-following natural language queries following the paradigm of InstructHumanEval.\footnote{\url{https://huggingface.co/datasets/codeparrot/instructhumaneval}} 

\colorbox{algblue}{\textcolor{white}{Algorithm}} \textbf{MBPP} \cite{austin2021program} includes 1,000 code generation queries. We randomly sample 150 instances from its sanitized test set \citep{evalplus}.

{\colorbox{mathblue}{Math}} \textbf{GSM8K} \citep{cobbe2021gsm8k} is a graduate-level math word problem dataset containing 8,790 instances. We randomly sample 150 instances from its test set. 

{\colorbox{mathblue}{Math}} \textbf{SVAMP} \citep{patel2021nlp}  is a collection of 1,000 elementary-school math problems. We randomly sample 150 instances from its test set.

\colorbox{logicblue}{\textcolor{white}{Logic}} \textbf{LogicBench} \citep{parmar2024logicbench}  comprises various logic questions in both binary classification and multiple-choice formats. We sample 100 binary and 100 multiple-choice questions, collecting 200 samples in total.

\colorbox{logicblue}{\textcolor{white}{Logic}} \textbf{Folio (original+perturbed)} \citep{han2022folio, wu2023reasoning} Original Folio is a manually curated logic benchmark. We select 81 instances along with their manually perturbed versions from \citet{wu2023reasoning}, yielding 162 instances.

{\colorbox{comprblue}{Integrated}} \textbf{AsyncHow} \citep{lin2024graph} is a planning reasoning benchmark. LLMs must interpret natural language descriptions (i.e., logic), find different possible paths in the graph (i.e., algorithm), and then calculate and compare the time cost for these paths (i.e., math) to reach the correct answer. We use stratified sampling based on the dataset’s complexity metrics and include 240 instances.

\subsection{Annotation and Quality Assurance}
\label{sec:annotation_quality_check}

After data sourcing, we hire AAVE speakers to rewrite each instance in AAVE. We schematize our annotation and validation in Figure~\ref{fig:annotation-redial} and describe them below, by which we ensure the consistency, representativeness, and neutrality of our dataset.

\paragraph{Annotation.} We hire and instruct AAVE speakers to rewrite each SE query so that it sounds natural to AAVE speakers while retaining all critical information (e.g., numerical values, logical conditions, and technical details). For algorithm-related tasks involving code, we hire annotators with a background in computer science to ensure the logic and semantics of the code tasks. In total, we hire 13 annotators with different demographic backgrounds to reduce personal biases.\footnote{Details on annotator compensation, qualifications, guidelines, and demographic distribution are presented in Appendix~\ref{sec:rubrics}.}

\paragraph{Validation.} We then perform a careful quality check to ensure both \textit{naturalness} and \textit{correctness}. First, we ask annotators to cross-check and edit each others' annotations to make sure that the annotations are \textbf{natural} to AAVE speakers. This can also further reduce individual annotator bias. Second, to ensure \textbf{correctness}, we first have non-AAVE speakers manually check the essential information, then conduct a sanity check with GPT-4o for the correctness of rewriting (details in Appendix~\ref{sec:data_verification}). We \textbf{manually check} data that GPT-4o flags as invalid to see if all essential information is preserved. \textbf{No instance is rejected solely based on the LLM's judgment.} We return invalid instances to AAVE speakers for correction and iterate the process until all data passes the check.

After these efforts, we obtain ReDial: a high-quality, end-to-end human-annotated SE–AAVE parallel dataset comprising over 1.2K instances spanning four canonical reasoning tasks. In the rest of this paper, we refer to the SE portion of the dataset as \textit{SE ReDial} and the AAVE portion as \textit{AAVE ReDial}.

\section{Experiments}
\label{sec:experiment}
\subsection{Experimental Setting}
\label{sec:setting}

\renewcommand{\arraystretch}{1.3}
\begin{table*}[t!]
\scriptsize
\centering
\resizebox{\textwidth}{!}{
\setlength{\tabcolsep}{2pt}
\begin{tabular}{l|l|cc|cc|cc|cc|cc|ccc}
\hline
\multirow{2}{*}{\textbf{Model}} & \multirow{2}{*}{\textbf{Setting}} & \multicolumn{2}{c|}{\textbf{Algorithm}} & \multicolumn{2}{c|}{\textbf{Logic}} & \multicolumn{2}{c|}{\textbf{Math}} & \multicolumn{2}{c|}{\textbf{Integrated}} & \multicolumn{3}{c}{\textbf{All}}\\
\cline{3-13}
 &  & SE & AAVE & SE & AAVE & SE & AAVE & SE & AAVE & SE & AAVE & \textbf{$\Delta$} \\
\hline
GPT-o1 & Direct & 0.818 & 0.825 & 0.947 & 0.923 & 0.878 & 0.815 & 0.942 & 0.925 & 0.892 & 0.866 & \textbf{-0.026} \\
\multirow{2}{*}{GPT-4o} & Direct & 0.790 & 0.761 & 0.933 & 0.930 & 0.818 & 0.768 & 0.783 & 0.312 & 0.832 & 0.716 & \textbf{-0.116} \\
& CoT & 0.771 & 0.761 & 0.950 & 0.920 & 0.815 & 0.771 & 0.762 & 0.662 & 0.826 & 0.784 & \textbf{-0.043} \\
\multirow{2}{*}{GPT-4} & Direct & 0.742 & 0.723 & 0.840 & 0.713 & 0.796 & 0.749 & 0.217 & 0.133 & 0.678 & 0.612 & \textbf{-0.067}\\
& CoT & 0.723 & 0.608 & 0.920 & 0.813 & 0.793 & 0.743 & 0.283 & 0.058 & 0.706 & 0.590 & \textbf{-0.115}  \\
\multirow{2}{*}{GPT-3.5-turbo} & Direct & 0.653 & 0.631 & 0.667 & 0.443 & 0.533 & 0.544 & 0.200 & 0.129 & 0.531 & 0.460 & \textbf{-0.072} \\
& CoT & 0.646 & 0.551 & 0.753 & 0.543 & 0.503 & 0.425 & 0.075 & 0.067 & 0.517 & 0.416 & \textbf{-0.101} \\
\midrule
\multirow{2}{*}{Claude-3.5-Sonnet} & Direct & 0.771 & 0.806 & 0.970 & 0.930 & 0.851 & 0.776 & 0.879 & 0.717 & 0.865 & 0.810 & \textbf{-0.055} \\
& CoT & 0.774 & 0.736 & 0.953 & 0.940 & 0.859 & 0.796 & 0.900 & 0.771 & 0.868 & 0.811 & \textbf{-0.058} \\
\midrule
\multirow{2}{*}{Llama-3.1-70B-Instruct} & Direct & 0.726 & 0.653 & 0.767 & 0.893 & 0.702 & 0.630 & 0.392 & 0.112 & 0.663 & 0.599 & \textbf{-0.064} \\
& CoT & 0.723 & 0.653 & 0.880 & 0.870 & 0.809 & 0.768 & 0.579 & 0.500 & 0.759 & 0.711 & \textbf{-0.049} \\
\multirow{2}{*}{Llama-3-70B-Instruct} & Direct & 0.682 & 0.643 & 0.907 & 0.887 & 0.663 & 0.552 & 0.158 & 0.067 & 0.628 & 0.562 & \textbf{-0.066} \\
& CoT & 0.697 & 0.646 & 0.923 & 0.887 & 0.616 & 0.561 & 0.517 & 0.350 & 0.693 & 0.622 & \textbf{-0.072} \\
\multirow{2}{*}{Llama-3-8B-Instruct} & Direct & 0.535 & 0.510 & 0.827 & 0.800 & 0.478 & 0.464 & 0.025 & 0.067 & 0.489 & 0.480 & -0.009\\
& CoT & 0.532 & 0.478 & 0.827 & 0.800 & 0.475 & 0.492 & 0.029 & 0.025 & 0.488 & 0.472 & -0.016 \\
\midrule
\multirow{2}{*}{Mixtral-8x7B-Instruct-v0.1} & Direct & 0.452 & 0.401 & 0.520 & 0.340 & 0.414 & 0.240 & 0.100 & 0.075 & 0.388 & 0.274 & \textbf{-0.114} \\
& CoT & 0.468 & 0.411 & 0.687 & 0.567 & 0.384 & 0.285 & 0.133 & 0.071 & 0.431 & 0.345 & \textbf{-0.086} \\
\multirow{2}{*}{Mistral-7B-Instruct-v0.3} & Direct & 0.331 & 0.255 & 0.400 & 0.213 & 0.315 & 0.271 & 0.096 & 0.075 & 0.297 & 0.214 & \textbf{-0.083} \\
& CoT & 0.312 & 0.245 & 0.453 & 0.347 & 0.323 & 0.293 & 0.083 & 0.083 & 0.305 & 0.252 & \textbf{-0.053} \\
\midrule
\multirow{2}{*}{Phi-3-Medium-Instruct} & Direct & 0.545 & 0.433 & 0.867 & 0.790 & 0.500 & 0.470 & 0.050 & 0.038 & 0.513 & 0.454 & \textbf{-0.059} \\
& CoT & 0.548 & 0.455 & 0.860 & 0.827 & 0.492 & 0.439 & 0.067 & 0.029 & 0.513 & 0.458 & \textbf{-0.055} \\
\multirow{2}{*}{Phi-3-Small-Instruct} & Direct & 0.615 & 0.252 & 0.820 & 0.760 & 0.530 & 0.525 & 0.058 & 0.062 & 0.530 & 0.421 & \textbf{-0.109}  \\
& CoT & 0.570 & 0.194 & 0.893 & 0.843 & 0.544 & 0.522 & 0.096 & 0.079 & 0.549 & 0.429 & \textbf{-0.119}\\
\multirow{2}{*}{Phi-3-Mini-Instruct} & Direct & 0.557 & 0.427 & 0.520 & 0.550 & 0.605 & 0.525 & 0.021 & 0.042 & 0.456 & 0.410 & \textbf{-0.046} \\
& CoT & 0.576 & 0.443 & 0.773 & 0.750 & 0.622 & 0.528 & 0.017 & 0.021 & 0.528 & 0.461 & \textbf{-0.067} \\

\hline
\end{tabular}
}
\caption{We report model pass rates using direct and CoT prompting on ReDial, including individual performances on subtasks and overall performance/gap (in column \textbf{all}). We follow the recommendations from \citet{dror2018hitchhiker} and
    test the statistical significance of performance differences between SE and AAVE on \textbf{all} results using the McNemar’s test for binary data \citep{mcnemar1947note}. We correct p-values for multiple measurements using the Holm-Bonferroni method \citep{holm1979simple}.  Statistically significant drops are in \textbf{bold}. Details in Appendix~\ref{sec:full_results}.}
    \label{tab:all_res_main}
\end{table*}
\subsubsection{Models}
\label{sec:models}

We test five families of models, two proprietary and three open-source. The rationale is to benchmark widely used LLMs with impressive reasoning performance. All experiments were conducted between September and December 2024.

\paragraph{GPT.} We use GPT-o1~\citep{OpenAIOS}, GPT-4o~\citep{Hurst2024GPT4oSC}, GPT-4~\citep{achiam2023gpt}, GPT-3.5-turbo~\citep{achiam2023gpt},\footnote{Proprietary model API version information: o1: GPT-o1-preview; gpt-4o: 2024-05-13; gpt-4: 2024-05-03; gpt-3.5-turbo: 2023-11-06.} 
as a family of closed-source models to compare with open-source models for dialect robustness. 
In particular, o1 is trained using large-scale reinforcement learning (RL) to reason through CoT and scales inference time computation to achieve highly complex reasoning paths, demonstrating significant improvements in reasoning tasks~\citep{OpenAIOS}.  We use the GPT-o1 model to understand how RL reasoning post-training affects LLMs' dialect robustness and fairness.

\paragraph{Claude.} Developed by Anthropic, the Claude 3 model family represents a widely-used proprietary LLM. For our experiments, we utilize the Claude 3.5 Sonnet model~\citep{TheC3}.

\paragraph{Llama.} We use Llama-3-8B / 70B-Instruct and Llama-3.1-70B-instruct \citep{dubey2024llama} which are reported for comparable performance with proprietary GPT models.

\paragraph{Mistral / Mixtral.} We use Mistral-7B-Instruct-v0.3 \citep{jiang2023mistral} and Mixtral-8x7B-Instruct-v0.1 \citep{jiang2024mixtral}. Mistral-7B-Instruct-v0.3 is reported to be outstanding in reasoning; with Mixtral-8x7B-Instruct-v0.1, we can understand whether Mixture-of-Expert architectures enhance dialect robustness.

\paragraph{Phi.} We use Phi-3-Mini / Small / Medium-128K-Instruct \citep{gunasekar2023textbooks, abdin2024phi} in our experiment. Phi-3 models, pre-trained on carefully designed ``textbook'' data, are reported for impressive performance in reasoning despite their small sizes (3.8/7/14B parameters each). We use these models to understand how highly curated pre-training data affect LLMs' dialect robustness and fairness.

\begin{figure*}[t!]
    \centering
    \includegraphics[trim=7 23 5 5, clip, width=\linewidth]{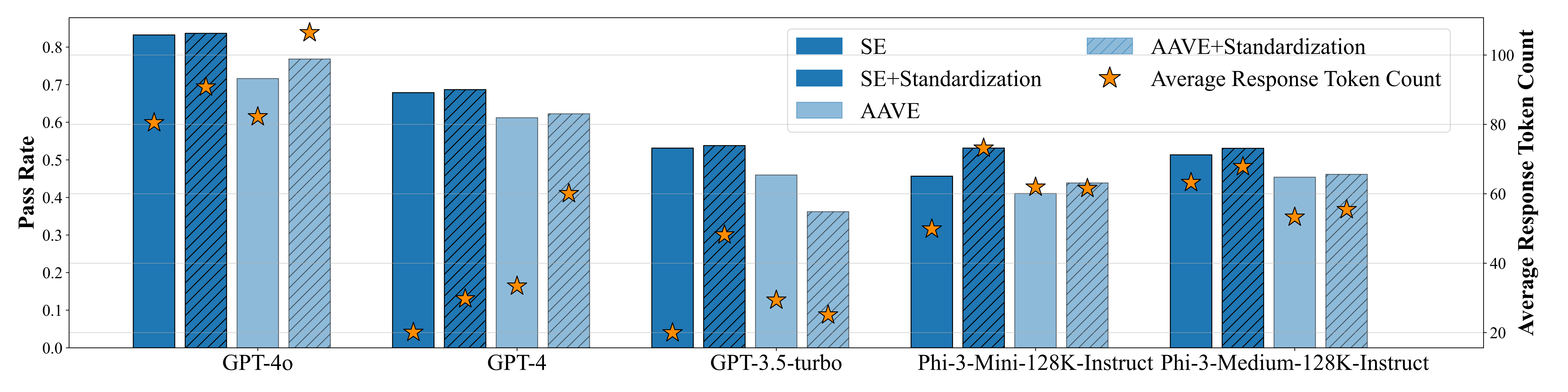}
    \caption{Model pass rate and average response token count before and after being prompted for standardization.  Standardization prompting generally improves LLM performance in both SE and AAVE ReDial (bar plot). However, even AAVE ReDial with standardization prompting cannot reach LLMs' vanilla performance in SE ReDial, even though they also tend to result in more tokens generated and thus higher inference cost (scatter plot).}
    \label{fig:think_sae_res}
\end{figure*}

\subsubsection{Implementation and Evaluation}
\label{sec:experiment_setting}
\paragraph{Implementation.} We set the temperature to zero for the main experiments to ensure maximum reproducibility. We report two prompting methods in our main results: (i) direct prompting LLMs with task instances, which resembles general real-life use cases the most (Direct) and (ii) zero-shot Chain of Thought \citep[CoT;][]{kojima2022large,wei2022chain}, i.e., adding instructions in the spirit of ``\texttt{Let's think step by step}'' on top of task descriptions, which resembles expert user prompts to improve model performance. For GPT-o1, we only test direct prompting due to its inherent CoT reasoning pattern.\footnote{We also test non-zero temperatures for a subset of the models and report results in Appendix~\ref{sec:non_zero_temperature}.}
We report further implementation details in Appendix~\ref{sec:implementation_details}.\footnote{We deliberately avoid testing advanced prompting methods, such as Tree of Thought \citep{yao2024tree} and Self-Refine \citep{madaan2024self}. Our focus is on evaluating \textbf{how LLMs perform when prompted for everyday use by dialect users}, which is critical for assessing fairness in LLMs. Similarly, we do not fine-tune any models, as our study aims to investigate biases inherent in models prior to task-specific adaptation. The effects of fine-tuning are beyond the scope of this study.}

\begin{table}[t]
    \centering
    \resizebox{\linewidth}{!}{
    \begin{tabular}{c|c|c|c|c|c}
    \toprule
        & \textbf{Algorithm} & \textbf{Math} & \textbf{Logic} & \textbf{Integrated} & \textbf{All} \\\midrule
    SE&0.632 & 0.622 & 0.768 & 0.302 & 0.597\\
AAVE&0.563 & 0.564 & 0.706 & 0.212 & 0.529\\
\midrule
$\Delta$&\textbf{-0.069} & \textbf{-0.058} & \textbf{-0.062} & \textbf{-0.090} & \textbf{-0.068}\\
\bottomrule
    \end{tabular}%
    }
    \caption{Pass rates by task averaged across responses from all models with direct prompting. In \textbf{bold}, results show statistically significant differences according to McNemar’s tests applied to AAVE and SE (i.e., models have significant drops in AAVE).  We also report the SE-AAVE absolute delta in performance.}
    \label{tab:perf_by_task}
    \vspace{-0.5cm}
\end{table}

\paragraph{Evaluation.} To unify evaluation metrics, we consider the pass rate for all tasks. For Algorithm, we consider Pass@1 using all base and extra unit test cases in EvalPlus~\citep{evalplus}, which results in either pass or fail for every code generation. 
We convert all other task measures of correctness or incorrectness to pass or fail.

\subsection{Experimental Results}
\label{sec:main_results}

\begin{figure*}[t!]
    \centering
    \includegraphics[trim=7 7 5 5, clip, width=\linewidth]{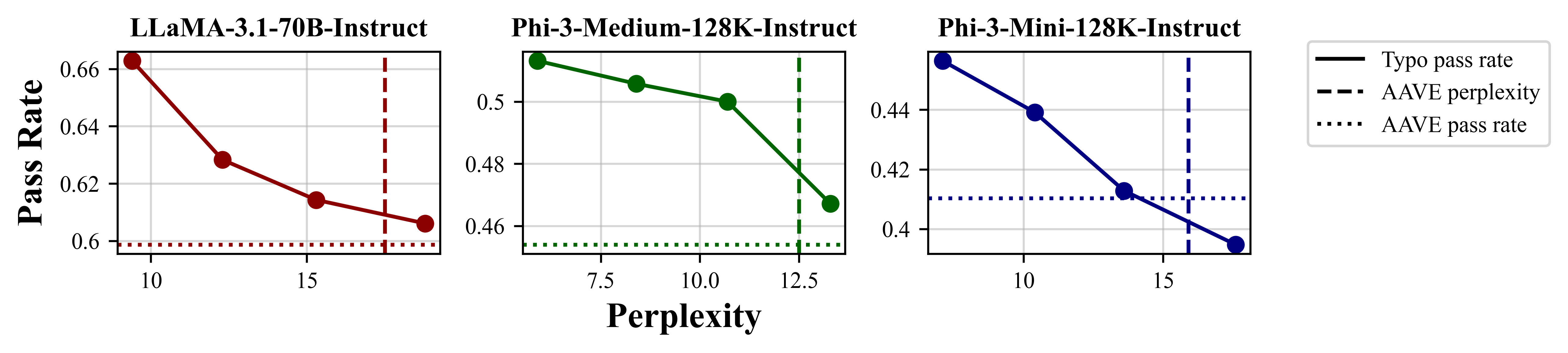}
    \caption{Model performance on misspelled SE compared to human-written AAVE data. We gradually add noise to SE ReDial to increase its perplexities until they surpass the perplexity of AAVE ReDial and report the models' performance on every perturbation level. Horizontal and vertical lines refer to model pass rates/perplexities on AAVE ReDial respectively. Larger LLMs (i.e., Llama-3.1-70B-Instruct and Phi-3-Medium-128K-Instruct) perform worse on AAVE than on perturbed text with a similar perplexity level.}
    \label{fig:noise_vs_aave}
    \vspace{-0.5cm}
\end{figure*}

We report pass rates for ReDial in Table~\ref{tab:all_res_main} and ~\ref{tab:perf_by_task},
and summarize the main results of our experiments.

\paragraph{All Models are Brittle.}  
All models experience performance drops in AAVE compared to SE ReDial, and these drops are statistically significant in all cases, with the sole exception of Llama-3-8B-Instruct. This indicates that our benchmark poses huge challenges to models, both in terms of absolute performance and with respect to their dialect robustness and fairness.

\textbf{The absolute performance gaps commonly range from around 5\% to over 10\%} ($\Delta$ in Table~\ref{tab:all_res_main}). Specifically, GPT-4o (zero-shot) shows an absolute gap of 11.6\%, dropping from an average of 0.832 to 0.716. GPT-4 (CoT) exhibits an 11.5\% drop.  Mixtral-8x7B-Instruct-v0.1 (zero-shot) shows a particularly large difference of 11.4\% points as well. Interestingly, we found that although the performance drop of GPT-o1 is smaller than other GPT models, it is still significant. This indicates that although further RL post-training on general reasoning and inference scaling can systematically enhance dialect robustness and fairness, they cannot completely solve the problem. 

In short, dialect unfairness and brittleness are identified in all the models we examined, including the mixture of expert and large reasoning models. This finding indicates that the problem is widespread, non-trivial, and cannot be easily mitigated by naively changing model architecture or proposing more complex reasoning paths.

\paragraph{All Tasks are Brittle.}  
When aggregated by task type, AAVE queries cause a statistically significant performance drop across all these categories (Table~\ref{tab:perf_by_task}). For instance, when averaging results across all models,  inputs written in AAVE (via direct prompting) lead to an average 10\% relative performance drop.

Interestingly, integrated reasoning tasks, which require multiple reasoning skill compositions, show some of the largest relative drops (about 30\%). This suggests that compositionally complex task may be more prone to dialect brittleness.

\paragraph{Prompting and Inference Scaling are Brittle.}  
While CoT prompting can slightly reduce the discrepancy for some models, it fails to close it entirely. For example, GPT-4o’s performance gap decreases from about 0.116 (zero-shot) to 0.043 (CoT). This suggests that even when models are given additional reasoning ``scaffolding,'' their understanding and performance in AAVE remain comparatively weaker than in SE, which is also in line with our observation with GPT-o1 results. We also try to bridge the gap by telling the LLMs to \textit{rephrase in Standard English then answer the question} (i.e., standardization), but this does not cancel the performance gap, while only introducing more inference cost (Figure~\ref{fig:think_sae_res}). \textbf{This means that even if dialect users pay more, they might still not receive the same quality service as users who use SE.}

\paragraph{Model Scaling is Brittle.} 
All model families display some degree of dialect-related performance degradation. A notable observation is that simply using larger models does not inherently improve robustness to AAVE. For example, even Llama-3.1-70B-Instruct, among the largest and most capable tested models, suffers from significant performance drops on AAVE queries. This pattern holds across the board, indicating that scaling model size alone is insufficient to address dialect-related performance disparities.

\section{Discussions}
\label{sec:analysis}
This section investigates the potential reasons for AAVE's brittleness. We show that LLMs' brittleness with AAVE reasoning queries is not simply due to the lack of understanding of this dialect or simple lexical features. The nuanced conversational norms of AAVE also contribute significantly to LLMs’ difficulties.

\subsection{General Understanding and Morphosyntactic Features}
\label{sec:ppl}
One possible explanation for the performance drop is that LLMs cannot process AAVE. We thus computed perplexities on ReDial AAVE vs.\ SE prompts in Llama-3.1-70B-Instruct, Phi-3-Medium-128K-Instruct, and Phi-3-Mini-128K-Instruct. Indeed, Table~\ref{tab:ppl_ReDial} confirms that LLMs exhibit higher perplexities on dialect than SE.

\begin{table}[t]
    \centering
    \resizebox{0.75\linewidth}{!}{%
    \begin{tabular}{c|c|c}
    \toprule
        Models & SE & AAVE \\\midrule
        Llama-3.1-70B-Instruct & 9.4 & 17.5 \\
        Phi-3-Medium-128K-Instruct & 5.9 & 12.5 \\
        Phi-3-Mini-128K-Instruct & 7.1 & 15.9 \\
        \bottomrule
    \end{tabular}%
    }
    \caption{Averaged perplexities across instances calculated by different models on SE/AAVE ReDial.}
    \label{tab:ppl_ReDial}
    \vspace{-0.5cm}
\end{table}

However, is the insufficient understanding the only reason for LLMs' performance to drop? To investigate this further, we gradually inject typos into SE ReDial by replacing/deleting/adding words and characters, such that we make the input texts more difficult for LLMs (i.e., the measured perplexity goes up). We present results in Figure~\ref{fig:noise_vs_aave}. While these perturbations degrade model performance, the drop does not reach the severity observed with natural AAVE data on large-scale models. This suggests that AAVE brittleness is not solely due to difficulties in text comprehension.

If language-agnostic processing ability cannot explain LLMs' brittleness, can we attribute the problem to morphosyntactic AAVE features? Following~\citet{ziems2022value, ziems-etal-2023-multi}, we use morphosyntactic transformation rules to inject AAVE features into SE ReDial. We find that performance degradation generally intensifies as the density of AAVE-specific features increases (see full results in Appendix~\ref{sec:multivalue_perturbation}). This suggests that these features play a significant role in diminishing model performance.

However, even under the most extreme synthetic perturbations, performance drops are notably less severe than those observed with human-rewritten prompts. This underscores \textbf{the critical importance of our high-quality human-annotated dialect data ReDial} for evaluating LLM fairness and robustness. Synthetic rule-based transformations provide valuable insights, yet fail to capture the contextual depth of real-world dialect usage. 

\subsection{AAVE Conversational Norms}
\label{sec:qualitative_analysis}
We use the mutual information between the token distributions of SE and AAVE ReDial to find that the top five most informative AAVE features in terms of distinguishing them from SE are \textit{'}, \textit{up}, \textit{in}, \textit{gon}, and \textit{gotta}. Note that many features here are not well-known AAVE-specific features (e.g., \textit{up}). Through a further investigation of our dataset, we find that these lexicons are associated with phrase-level AAVE constructions. For instance, instead of saying \textit{...encode the answer...} in SE, AAVE instruction says \textit{...wrap it up...}. This finding is particularly interesting because, in addition to previous linguistic observations of AAVE morphosyntactic features \cite{sidnell2002african, martin2013sentence}, there are important conversational norms of the dialect, such as nuanced uses of phrases \cite{green2002african, morgan2002language, rickford2007spoken}.

We compute Spearman's correlation between the frequency of the features we find with mutual information in each instance and their corresponding performance drop. Indeed, these features play a significant role in predicting GPT-4o's performance degradation (\textit{r=-0.318}, \textit{p<0.001}). We further implement and analyze 12 rule-based AAVE features following \citet{ziems2022value} (details in Appendix~\ref{sec:synthetic_lexical_feature}), which are well documented in linguistic literature such as \textit{finna} as a marker of immediate future \cite{nguyen-grieve-2020-word}. We notice that the influential lexical features are a subset of the feature set discovered by mutual information (i.e., some of the actual influential features are not encoded in synthetic transformation rules). Consequently, the influence of synthetic features is not as strong as those discovered by mutual information (\textit{r=-0.256}, \textit{p<0.001}). This means that simple rule-based transformations that implement the most salient morphosyntactic AAVE features may not be able to capture rich, context-dependent use of the dialect and, therefore, fall short in predicting LLMs' performance in real workflows.

With assistance from GPT-o1 preview to filter the vast amount of data, we conducted a linguistic analysis of frequent errors in AAVE. 
For algorithm tasks, grammatical constructions and non-standard verb forms (e.g., \textit{finna'}, \textit{'em}), omission of articles and auxiliary verbs may cause the model to misinterpret references and function naming conventions. For example, GPT-4o interprets \textit{you gon' write a python function, python\_function} as a general statement rather than a directive to name the function.
On logic tasks, the frequent use of double negatives, zero copula, and inverted conditionals introduces structural ambiguities (e.g., \textit{He don’t take no breaks}). 
On math, informal expressions, unclear quantity references, and non-standard comparative constructions cause erroneous parsing of numerical information and confusion over collective versus individual quantities. 
Informal phrasings like \textit{half as much as he be runnin'} and ambiguous comparative expressions (\textit{4 fewer boxes of apple pie than on Sunday}) can cause the model to misinterpret numerical relationships.
On the integrated task, phonetic spellings, colloquial connectors, and inverted word orders limit the model’s ability to understand concurrency and follow stepwise instructions.  
Such dialectal nuances highlight the necessity of our dataset and also call for more efforts to collect more human data for relevant purposes.\footnote{We also provide more analysis comparing reasoning chains in AAVE/SE ReDial in Section~\ref{sec:standardized_prompting_qual_analysis}.}

\section{Related Work}
\label{sec:related_works}
\paragraph{Dialect studies in natural language processing.}
Previous work on AAVE studies in natural language processing mostly focuses on non-reasoning-heavy tasks such as POS tagging \citep{jorgensen-etal-2015-challenges,jorgensen-etal-2016-learning}, language identification and dependency parsing \citep{blodgett-etal-2016-demographic}, automatic captioning \citep{tatman-2017-gender}, and language modeling \citep{deas-etal-2023-evaluation}.
AAVE is found to be more likely to trigger false positives in hate speech identifiers \citep{davidson2019racial, sap-etal-2019-risk} due to word choices \citep{harris2022exploring}, to be considered negative by automatic sentiment classifier \citep{groenwold-etal-2020-investigating}, and to cause covert biases in essential areas of social justice \citep{hofmann2024dialect}. Other studies \citep{ziems2022value, gupta2024aavenue} find that rule-based AAVE perturbations can downgrade language model performance in GLUE \citep{wang2018glue}.

More generally, dialects across world languages pose challenges to natural language processing systems. \citet{ziems-etal-2023-multi} find that auto-encoder models are brittle on rule-based English dialect feature perturbations. \citet{fleisig2024linguistic} report that responses generated by chatbots to dialectal inputs are perceived as more negative by English dialect speakers than responses to SE prompts. \citet{faisal2024dialectbench} find that non-standardized dialects cause problems in dependency parsing \citep{scherrer2019digitising} and machine translation \citep{mirzakhalov2021turkic} on mBERT and XLM-R \citep{conneau-etal-2020-unsupervised}.

\textbf{Fairness and Robustness of Large Language Models}. LLMs exhibit both unfairness and brittleness. They offer unfair performance \citep{huang2023not, dong2024evaluating} and cost burdens \citep{petrov2024language} to users of different languages, marginalize minority groups across dimensions such as gender \citep{kotek2023gender, fraser2024examining}, race \citep{hofmann2024dialect, wang2024large,sun-etal-2025-aligned}, and culture \citep{naous2023having, tao2024cultural}. They also show different performance to reasoning tasks in different languages (\citealt{huang2023not, huang2024mindmerger, ranaldi2024empowering}; inter alia). To our knowledge, our study provides the first extensive empirical evidence that LLMs are unfair in reasoning tasks. This bias specifically affects speakers of certain dialects within a single language.

Previous works report that LLMs are brittle when prompts are varied through typos or paraphrasing in SE \citep{elazar-etal-2021-measuring,liang2022holistic, raj2022measuring, zhu2023promptbench, lin2024graph,rottger-etal-2024-political}. We use a novel approach of human-written perturbations in AAVE and evaluate LLM robustness towards these natural perturbations, which result in greater brittleness than synthetic typo-style (Section~\ref{sec:ppl}) or linguistic-rule-based (Appendix~\ref{sec:multivalue_perturbation}) perturbations.

\section{Conclusion}
\label{sec:conclusion}
Our study is the first to objectively evaluate the dialect robustness and fairness of LLMs in reasoning. We introduce \textbf{ReDial}, a dataset of over 1.2K parallel prompts in Standardized English and African American Vernacular English (AAVE) tailored to algorithm, logic, math, and integrated reasoning. Extensive empirical evidence on ReDial demonstrates that LLMs exhibit significant unfairness and brittleness when reasoning tasks are expressed in AAVE. These findings underscore the unfairness to dialect users and LLMs' brittleness with natural prompt variations with the same semantics. We advocate for further research to enhance dialect fairness and robustness of LLMs, ensuring equal service for all linguistic groups and demographics.

\section{Limitations}
\label{sec:limitation}

First, as the first systematic framework for analyzing LLM bias in dialectal queries for reasoning tasks, we selected AAVE due to its linguistic significance and cultural impact. However, we recognize the vast diversity of dialects worldwide. The insights derived from AAVE may not generalize to other dialects. To ensure annotation quality and maintain the focus of our study, we concentrated on AAVE with high-quality human annotations. Also, we only have one annotator to annotate each instance and another AAVE annotator for naturalness check due to the limited budget. While we try to ensure diversity by hiring more annotators (13 vs.\ 3 in previous literature; \citealp{ziems2022value}) and spreading annotation over more instances, we are aware that this might still bring  subjectivity into our benchmark. Future research could expand on our framework to encompass a wider range of dialects, hiring a more diverse range of annotators, and generating more broadly applicable conclusions.

Second, our benchmark, ReDial, evaluates LLM performance across four categories of reasoning tasks using queries sampled from seven popular and well-documented benchmarks. While these tasks are representative of common reasoning challenges in both fundamental (e.g., the logic tasks) and practical areas (e.g., code generation and complex planning for multi-agent systems in the integrated tasks), we acknowledge that reasoning is a multifaceted domain with many additional categories and tasks that fall outside the scope of this study (e.g., medical/financial reasoning).

Third, we evaluated five representative LLM families in this study, including widely used and state-of-the-art models. However, given the rapid proliferation of new LLMs, testing every model is infeasible. We hope that future research will use the ReDial benchmark to investigate fairness and reasoning robustness across a broader range of LLMs as they emerge.

Fourth, due to the difficulty of gathering large-scale dialect data for training, we cannot perform additional analysis on how supervised fine-tuning/reinforcement learning with relevant data might help models bridge the dialect gap, which makes it difficult to draw conclusions about how different training methods affect dialect robustness. Although it might be technically possible to generate high-quality synthetic data (with known difficulties discussed in Sections~\ref{sec:introduction} and~\ref{sec:analysis}), we consider it to be out of the scope of our paper. Despite so, we do observe that general supervised fine-tuning/reinforcement learning do not bridge the gap as can be observed in comparing models in the GPT family. We hope that future research can develop a reliable and scalable way to gather more high-quality dialect data.

Last but not least, while we present extensive empirical evidence demonstrating the performance drop of LLMs on dialectal queries, our study does not deeply investigate the underlying causes of these performance discrepancies or propose systematic methods to mitigate this bias. These topics exceed the scope of our work but are critical for addressing the inequities we have identified. Despite this limitation, we believe that ReDial provides a robust and systematic tool to help researchers explore these issues. The absence of immediate solutions should not detract from the significance of our findings, which lay the groundwork for future efforts to address fairness and robustness in LLMs.

\section{Ethic Statement}
\label{sec:ethic_statement}
ReDial is a collection of high-quality human-annotated translations: obtaining such data requires making clear design choices and poses ethical questions that we hereby address. 

For data collection, we deliberately do not set hard constraints for annotator identity and demographic verification, recognizing there are no definite boundaries to identify dialects and their speakers~\citep{king2020african}. 
The authors further elaborate that the term ``AAVE'' itself is contested, with alternatives that could be used instead; in employing the term ``AAVE'', we adhere to the widely used terminology in related works on dialects and NLP~\citep{ziems2022value,gupta2024aavenue}.
We corroborate the data quality by asking self-identified dialect speakers to cross-validate each others' answers. 

We do not force annotators to disclose their personal information; 
while we firmly commit to this rule to protect annotators' privacy, it makes it difficult to draw detailed conclusions about how annotators' backgrounds shape their writing/individual-level variations.
Further on the ethical aspect of data collection, we work with a data vendor that makes sure the recruitment and annotation adhere to high standards for and from the annotators. 
However, although we have a legal contract and we try our best to convey our guidelines and requirements, we admit that we do not have full control over how the vendor recruits people and conducts data annotation.

We also stress that the LLM validation stage in our quality control process is not completely trustworthy as even they are prone to hallucinations~\citep{ji2023survey} and biases against minority groups~\citep{xu2021detoxifying, fleisig2024linguistic, smith2024standard, wang2024large}. 
To mitigate this issue, we conduct full manual checks of every instance identified as invalid by an LLM so that no instance is rejected purely because of LLM decisions.

\section{Acknowledgement}
We thank all the bodies who have provided funding for the authors and for the associated project, including the OpenAI Research Access credits to FL. FL is supported by Clarendon and Jason Hu studentship. ELM is supported by the Alan Turing Institute. We thank Ambrosio Blanco for helpful feedback on the ethical review. We are grateful to the people who offered feedback and suggestions along the way, and in particular, Su Lin Blodgett, Wenshan Wu, Xiaoyuan Yi, Yan Xia, Jing Yao, Sunayana Sitaram, and other colleagues in Microsoft Research, who offered invaluable advice and helped us refine the paper.

\bibliography{custom}
\clearpage
\newpage
\appendix
\section{Appendix}
\subsection{Source Dataset Illustration}
\label{sec:per_dataset_illustration_appendix}
\subsubsection{Algorithm}
\begin{tcolorbox}
\colorbox{pink}{\textbf{Original HumanEval}}\newline

  \begin{lstlisting}[language=Python, basicstyle=\ttfamily\footnotesize, keywordstyle=\color{blue}, commentstyle=\color{green!60!black}]
from typing import List


def has_close_elements(numbers: 
List[float], threshold: float) 
-> bool:
    """ Check if in given list of 
    numbers, are any two numbers 
    closer to each other than 
    given threshold.
    >>> has_close_elements([1.0, 
    2.0, 3.0], 0.5)
    False
    >>> has_close_elements([1.0, 
    2.8, 3.0, 4.0, 5.0, 2.0], 0.3)
    True
    """
\end{lstlisting}

\colorbox{pink}{\textbf{InstructHumanEval Used in the Paper}}\newline

Write a function has\_close\_elements(numbers: List[float], threshold: float) -$>$ bool to solve the following problem: 

Check if in given list of numbers, are any two numbers closer to each other than given threshold.

$>>>$ has\_close\_elements([1.0, 2.0, 3.0], 0.5)

False

$>>>$ has\_close\_elements([1.0, 2.8, 3.0, 4.0, 5.0, 2.0], 0.3)

True 

\end{tcolorbox}

\begin{tcolorbox}
\colorbox{pink}{\textbf{MBPP}}\newline

Write a python function to remove first and last occurrence of a given character from the string.

Your code should pass these tests:

assert remove\_Occ(``hello",``l") == ``heo"

assert remove\_Occ(``abcda",``a") == ``bcd"

assert remove\_Occ(``PHP",``P") == ``H"

\end{tcolorbox}

\subsubsection{Logic}
\begin{tcolorbox}
\colorbox{pink}{\textbf{LogicBench}}\newline

If an individual consumes a significant amount of water, they will experience a state of hydration. Conversely, if excessive amounts of sugar are ingested, a sugar crash will ensue. It is known that at least one of the following statements is true: either the Jane consumes ample water or she will not experience a sugar crash. However, the actual veracity of either statement remains ambiguous, as it could be the case that only the first statement is true, only the second statement is true, or both statements are true.

Can we say at least one of the following must always be true? (a) she will feel hydrated and (b) she doesn't eat too much sugar
\end{tcolorbox}

\begin{tcolorbox}
\colorbox{pink}{\textbf{Folio}}\newline

Consider the following premises: ``People in this club who perform in school talent shows often attend and are very engaged with school events. People in this club either perform in school talent shows often or are inactive and disinterested community members. People in this club who chaperone high school dances are not students who attend the school. All people in this club who are inactive and disinterested members of their community chaperone high school dances. All young children and teenagers in this club who wish to further their academic careers and educational opportunities are students who attend the school. Bonnie is in this club and she either both attends and is very engaged with school events and is a student who attends the school or is not someone who both attends and is very engaged with school events and is not a student who attends the school."

Assuming no other commonsense or world knowledge, is the sentence ``Bonnie performs in school talent shows often." necessarily true, necessarily false, or neither? Answer either ``necessarily true", ``necessarily false", or ``neither".
\end{tcolorbox}

\subsubsection{Math}
\begin{tcolorbox}
    \colorbox{pink}{\textbf{GSM8K}}\newline

    Given a mathematics problem, determine the answer. Simplify your answer as much as possible and encode the final answer in $<$answer$><$/answer$>$ (e.g., $<$answer$>$1$<$/answer$>$).
    
    Question: Janet's ducks lay 16 eggs per day. She eats three for breakfast every morning and bakes muffins for her friends every day with four. She sells the remainder at the farmers' market daily for \$2 per fresh duck egg. How much in dollars does she make every day at the farmers' market?
    
    Answer:
\end{tcolorbox}

\begin{tcolorbox}
    \colorbox{pink}{\textbf{SVAMP}}\newline

    Given a mathematics problem, determine the answer. Simplify your answer as much as possible and encode the final answer in $<$answer$><$/answer$>$ (e.g., $<$answer$>$1$<$/answer$>$).
    
    Question: Winter is almost here and most animals are migrating to warmer countries. There are 41 bird families living near the mountain. If 35 bird families flew away to asia and 62 bird families flew away to africa How many more bird families flew away to africa than those that flew away to asia?
    
    Answer:
\end{tcolorbox}

\subsubsection{Comprehensive}

\begin{tcolorbox}
    \colorbox{pink}{\textbf{AsyncHow}}\newline

    To create a video game, here are the steps and the times needed for each step.
    
    Step 1. Learn the basics of programming (180 days)
    
    Step 2. Learn to use a language that is used in games (60 days)
    
    Step 3. Learn to use an existing game engine (30 days)
    
    Step 4. Program the game (90 days)
    
    Step 5. Test the game (30 days)\newline\newline    
    
    These ordering constraints need to be obeyed when executing above steps:
    
    Before starting step 2, complete step 1.
    
    Before starting step 3, complete step 1.
    
    Before starting step 4, complete step 2.
    
    Before starting step 4, complete step 3.
    
    Before starting step 5, complete step 4.\newline\newline
    
    Question: Assume that you need to execute all the steps to complete the task and that infinite resources are available. What is the shortest possible time to create a video game? Answer the time in double quotes.
    
    Answer:
\end{tcolorbox}

\subsection{ReDial Samples}
\label{sec:redial_samples}
\begin{tcolorbox}
\colorbox{pink}{\textbf{Algorithm}}\newline
\colorbox{lightgrey}{Standardized}

Write a function python\_function(numbers: List[float], threshold: float) $->$ bool to realize the following functionality:

Check if in given list of numbers, are any two numbers closer to each other than given threshold.

$>>>$ python\_function([1.0, 2.0, 3.0], 0.5)

False

$>>>$ python\_function([1.0, 2.8, 3.0, 4.0, 5.0, 2.0], 0.3)

True 

Generate a Python function to solve this problem. Ensure the generated function is named as python\_function.
\end{tcolorbox}

\begin{tcolorbox}
\colorbox{pink}{\textbf{Algorithm}}\newline
\colorbox{lightgrey}{AAVE}

Aight, so here you gonna write a function called python\_function(numbers: List[float], threshold: float) $->$ bool that gon' do this following functionality:

Aight, Listen. Say you got a list of numbers yeah? Now, we trynna see if any two of 'em numbers is closer to each other than a number you give, feel me?So, this is what we 'bout to do: 

$>>>$ python\_function([1.0, 2.0, 3.0], 0.5)

False

That's gon' give you False cuz ain't none of 'em numbers close enough.But, if you hit it like:

$>>>$ python\_function([1.0, 2.8, 3.0, 4.0, 5.0, 2.0], 0.3)

True

Bet you gettin' True, cuz this time some of 'em numbers real tight.

You gotta whip up a Python function to handle this problem. You gon' make sure the function name right, which gotta python\_function.
\end{tcolorbox}

\begin{tcolorbox}
\colorbox{pink}{\textbf{Math}}\newline
\colorbox{lightgrey}{Standardized}

Given a mathematics problem, determine the answer. Simplify your answer as much as possible and encode the final answer in $<answer></answer>$ (e.g., $<answer>1</answer>$).

Question: John is raising money for a school trip. He has applied for help from the school, which has decided to cover half the cost of the trip. How much money is John missing if he has \$50 and the trip costs \$300?

Answer:\newline
\end{tcolorbox}

\begin{tcolorbox}
\colorbox{pink}{\textbf{Math}}\newline
\colorbox{lightgrey}{AAVE}

Bet, so here's whatsup. Youn finna get a math problem, and you gon' tryna find the answer out. You gotta simplify that answer as much as possible tehn wrap it up inside $<answer></answer>$ (somethin' like this:, $<answer>1</answer>$).

Question: John been raisin' money fo' a school trip. He done ask the school fo' help, and they decided they gon' be coverin' half the trip cost. How much money John be missin' if he got \$50, and the trip cost \$300.

Answer:

\end{tcolorbox}

\begin{tcolorbox}
\colorbox{pink}{\textbf{Logic}}\newline
\colorbox{lightgrey}{Standardized}

Consider the following premises: "All bears in zoos are not wild. 

Some bears are in zoos. "

Assuming no other commonsense or world knowledge, is the sentence "Not all bears are wild." necessarily true, necessarily false, or neither? Answer either "necessarily true", "necessarily false", or "neither". Encode the final answer in $<answer></answer>$ (e.g., $<answer>$necessarily true$</answer>$).\newline
\end{tcolorbox}

\begin{tcolorbox}
\colorbox{pink}{\textbf{Logic}}\newline
\colorbox{lightgrey}{AAVE}

Aight, check this. You got 'em premises right here: "All bears in zoos ain't considered wild. 

There are some bears livin' in zoos. "

Ain't no using no other commonsense or world knowledge, you gon' try find out if the sentence "Not every bear out there be wild." necessarily true, necessarily false, or neither? Pick either "necessarily true", "necessarily false", or "neither". Then wrap that answer up in $<answer></answer>$ (e.g., $<answer>$necessarily true$</answer>$).
\end{tcolorbox}

\begin{tcolorbox}
\colorbox{pink}{\textbf{Comprehensive}}\newline
\colorbox{lightgrey}{Standardized}

To try fishing for the first time, here are the steps and the times needed for each step

Step 1. drive to the outdoor store (10 minutes)

Step 2.compare fishing poles (30 minutes)

Step 3. buy a fishing pole (5 minutes)

Step 4. buy some bait (5 minutes)

Step 5. drive to a lake (20 minutes)

Step 6. rent a small boat (15 minutes)\newline

These ordering constraints need to be obeyed when executing above steps:

Step 1 must precede step 2.

Step 2 must precede step 3.

Step 2 must precede step 4.

Step 3 must precede step 5.

Step 4 must precede step 5

Step 5 must precede step 6.\newline

Question: Assume that you need to execute all the steps to complete the task and that infinite resources are available. What is the shortest possible time to complete this task? What is the shortest possible time to complete this task? Encode the final answer in $<answer></answer>$ (e.g., $<answer>$1 min$</answer>$).

Answer:\newline
\end{tcolorbox}

\begin{tcolorbox}
\colorbox{pink}{\textbf{Comprehensive}}\newline
\colorbox{lightgrey}{AAVE}

If you finna go fish for the first time, here's what you got to know and the times you need for each step.

Step 1. To kick things off, pull up to the outdoor store (10 minutes)

Step 2. Check out which one of them fishing poles is good and which one is not (30 minutes)

Step 3. Cop a fishing pole (5 minutes)

Step 4.Get yourself some bait as well (5 minutes)

Step 5. Head out to a lake (20 minutes)

Step 6.rent yourself a small boat (15 minutes)\newline

These ordering constraints gotta be followed when you doin' 'em steps above:
You gotta deal with 1 before hittin' the 2.

You gotta deal with 2 before hittin' the 3.

You gotta deal with 2 before hittin' the 4.

You gotta deal with 3 before hittin' the 5.

You gotta deal with 4 before hittin' the 5.

You gotta deal with 5 before hittin' the 6.\newline

Question: Assumin' you outta do all 'em steps to finish up the task, and you got infinite resources. What the shortest time be to knock this task out? Wrap that answer up in $<answer></answer>$ (e.g., $<answer>$1 min$</answer>$).

Answer:
\end{tcolorbox}

\subsection{Rubrics}
\label{sec:rubrics}
\subsubsection{Employment Information}
We work with data vendors to employ 13 annotators in total for our task. For algorithm instance annotation, we specifically hire annotators with computer science backgrounds. Annotators are self-identified as proficient speakers of African American Vernacular English. We do not pose any hard constraints in verifying dialect identity as previous studies do (e.g., \cite{ziems-etal-2023-multi}). We note even within a dialect there can be significant variations on the individual level, and that we want to avoid homogenization and over-simplification of the dialect \citep{king2020african}. Instead, we ask self-identified annotators to cross-check each other's annotations and modify if they sound unnatural.

We only have one annotator for each prompt due to the budget limitation. However, we ensure diversity and reduce subjectivity in two ways for each annotation. First, we ensure diversity by hiring more annotators to annotate more instances (as opposed to one annotator per instance, which may introduce biases). We report the average score across datasets; this allows us to provide a diverse range of reference points, as the overall diversity is preserved by spreading annotations across many instances. For example, having 1,600 instances annotated by 4 annotators (with each annotator covering 400 instances) achieves a similar level of diversity as having 400 instances where all 4 annotators annotate every instance. After annotating for each instance, another AAVE speaker, who was not the annotator for that instance, ensures the resulting translation is consistent and sounds natural.

Details of employment are shown below.

\textbf{Information Collected} We do not force disclosure of personal information from our annotators (e.g., name, age, etc). We only make it mandatory that we collect the annotators' responses to our consent form and their annotations of our data.

\textbf{Demographic information} We report information on those (11 annotators) willing to disclose more demographic information. Annotators’ ages range between 23 and 35 years old, with 2 female and 9 male annotators. 4 of them have Master's degrees, and others have Bachelor's degrees.

\textbf{Risk and Consent} We note that our base datasets are from publicly available, widely used, peer-reviewed datasets that adhere to peer-review regulations. Moreover, our tasks are mainly centered around reasoning, which does not concern sensitive information per se. In addition, we make sure that annotators understand the risks of the annotation (i.e., although we have tried our best to ensure the safety of the data, it is still possible that they may feel uncomfortable in the annotation) and their right to exit the task during the process by signing a consent form prior to the start of the task.

\textbf{Compensation} We offer payment to annotators with hourly rates higher than the U.S. federal minimum wage.

\textbf{No AI Assistant} We explicitly inform our annotators that they should not reply on any AI assistant tools to help them complete the task. To further ensure this, we design our annotation platform to disallow copy and paste. The default annotation area for annotators is the original text, which means that it is easier for annotators to simply edit the text than to query AI assistants.

\subsubsection{Annotation Guideline}
You need to translate/rephrase/localize the task input in a way that is natural to the speakers of your dialect without changing the intention of the prompts. You should not change named entities, numbers, equations, variable names and other formal devices that are not natural language per se or those that would affect the intention of the prompts. The translation does not need to be grammatical or acceptable in standard English. Rather, it should accurately reflect the features of their dialects. You can add or delete some functional content to make the prompts sound more natural (e.g., adding fillers). However, you should keep the vital information complete and unchanged.  

You should NOT change information that would invalidate the output given the question. If you are unsure about any specific parts, leave them unchanged. Especially, you should not change the following parts: 

(i) numbers (e.g. 180 in 180 days) 

(ii) units (e.g. days in 180 days) 

(iii) equations and symbols (e.g., \textbackslash$[$f(x) = \textbackslash left \textbackslash\{ \textbackslash begin\{array\}\{cl\} ax+3, \& \textbackslash text\{ if \}$x>2$ in Let \textbackslash $[f(x) =$ \textbackslash left \textbackslash \{ \textbackslash begin\{array\}\{cl\} $ax+3$, \& \textbackslash text\{ if \}$x>2$) 

(iv) proper nouns (e.g., Natalia in Natalia sold clips to 48 of her friends) 

(v) function names, variables, data types, and input-output examples (e.g., $>>>$ has\_close\_elements($[1.0, 2.0, 3.0]$, 0.5) False $>>>$ has\_close\_elements($[1.0, 2.8, 3.0, 4.0, 5.0, 2.0]$, 0.3) True in Check if in given list of numbers, are any two numbers closer to each other than given threshold. $>>>$ has\_close\_elements($[1.0, 2.0, 3.0]$, 0.5) False $>>>$ has\_close\_elements($[1.0, 2.8, 3.0, 4.0, 5.0, 2.0]$, 0.3) True) 

\subsection{Data Quality Verification}
\label{sec:data_verification}
After we conduct human validations for \textit{naturalness} and \textit{correctness} of prompts, we conduct the final round sanity check with GPT-4o. We prompt GPT-4o with temperature 0.7 and sample three instances for each query. We manually inspect instances again where all of the answers suggest that they are invalid paraphrases of the original prompts.

\begin{tcolorbox}
\colorbox{white}{\textbf{User prompt}}\newline
You will be given two prompts, one in Standard English and one in African American English. Determine whether the African American English prompt is a valid paraphrase of the Standard English prompt. Ignore the semantic validaty of the Standard English prompt.

Standard English: "[SE\_PROMPT]"

African American English: "[AAVE\_PROMPT]"

Is the African American English prompt a valid paraphrase of the Standard English prompt?

\end{tcolorbox}

\subsection{Implementation Details}
\label{sec:implementation_details}
\subsubsection{Dataset Implementation}
For Algorithm, we unify the prompts by substituting all function names as python\_function to avoid as much memorization as possible. We also manually corrected instances in HumanEval where the task descriptions were not precise enough (e.g., when the output data structure specified in the docstring is different from the one specified in the function heading). We also slightly modified some instructions in algorithm datasets without changing their intention to make sure our prompts are coherent (e.g., changing \textit{to solve the following problem} to \textit{to realize the following functionality}).

For other tasks, we unify the task output by asking LLMs to encode answers in $<answer></answer>$ to enable easy parsing. All details can be found in ReDial dataset files.

\subsubsection{Inference Implementation}
We set temperature=0 and max new token as 4096 for all models at inference time unless specified in the main paper. We run experiments on GPT-4o/4/3.5 via Azure OpenAI service. We evaluate all other models via Azure Machine Learning Studio API for main results. Experiments run in the analysis part are hosted on 4 A100 with 80GB memory each.

\subsection{Results for Non-zero Temperature}
\label{sec:non_zero_temperature}
\begin{figure}[ht]
    \centering
    \includegraphics[width=\linewidth]{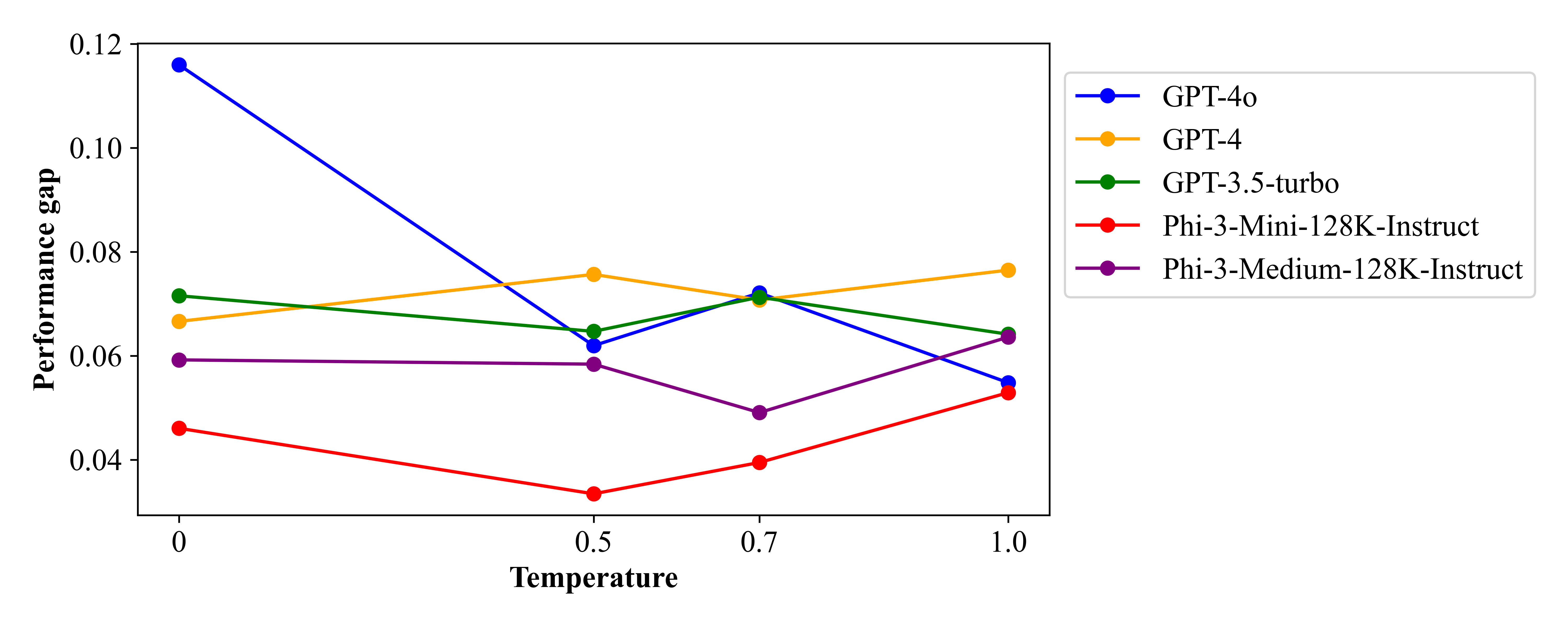}
    \caption{We vary the temperature by 0, 0.5, 0.7, 1 and report the performance gap between Standardized and AAVE ReDial.}
    \label{fig:vary_temperature}
\end{figure}

We vary the temperature by 0, 0.5, 0.7, and 1 on GPT-4o/4/3.5-turbo and Phi-3-Mini/Medium-128K-Instruct. When the temperature is not 0, we sample 3 answers per query and take average pass rates as results for corresponding settings. Results are in Figure~\ref{fig:vary_temperature}.

We find that increasing temperature reduces the gap for GPT-4o in general, but does not affect other models' performance as much. Even when the performance gap is reduced, increasing temperature cannot cancel the gap.

\subsection{Multivalue Perturbation}
\label{sec:multivalue_perturbation}
Since the unfamiliarity of data cannot explain the whole picture, how much can we attribute the failure to AAVE-specific features? We use the rule-based transformation method in~\cite{ziems-etal-2023-multi} to inject AAVE features into our dataset for synthetic probing. 
We compare GPT-4o/4/3.5 and Phi-3-Medium/Mini-128k-Instruct performance in feature densities of $\{0, 0.25, 0.5, 0.75, 1\}$ and run the same setting as the main experiment.

\begin{figure*}[ht]
    \centering    \includegraphics[width=0.8\linewidth]{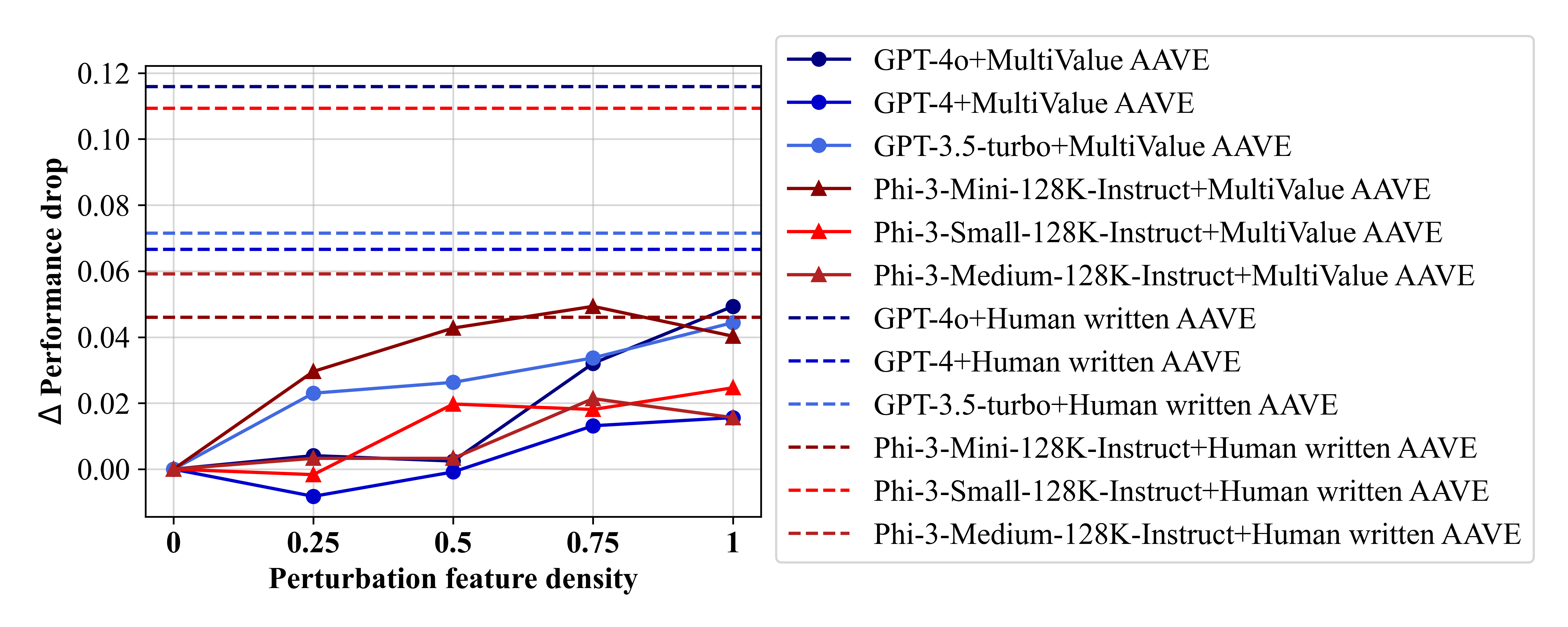}
    \caption{Perturbation with AAVE features. We control perturbation feature densities at $\{0, 0.25, 0.5, 0.75, 1\}$ to gradually inject AAVE features using rule-based transformations.}
    \label{fig:multivalue_perturb_vs_acc}
\end{figure*}

Results are shown in Figure~\ref{fig:multivalue_perturb_vs_acc}. On the one hand, we find that models generally show increasing performance drops with increasing feature density, which means that AAVE-specific features do contribute to model performance drops. On the other hand, even drops caused by the strongest perturbation are generally far from the drops caused by human-rewritten prompts. This shows the limitation of previous methods in revealing LLM robustness based on synthetic data as there can be more influential factors than what lexico-syntactic rules can capture. Phi-3-Mini-128K-Instruct is again an outlier here, being that it is the only model that has a stronger performance drop in feature injections compared to human-written dialect data.

\subsection{Qualitative Analysis of Reasoning Chains}
\label{sec:standardized_prompting_qual_analysis}
We qualitatively compared GPT-4o’s reasoning chain in SE and AAVE ReDial. We focus on the math subset of ReDial and identify two key error patterns: (1) distraction by irrelevant information and (2) failure to execute all steps. Below, we briefly sketch our findings.

\textbf{Distraction by irrelevant information.} GPT-4o gets distracted by task-irrelevant information in AAVE ReDial, while we do not observe the same in SE ReDial. For instance, in ‘Say we got 8 different books and 10 different movies in the crazy silly school series. How many more movies than books is there gon be in the crazy silly school series if you read 19 books and watched 61 movies?’, books that have been read and movies that have been watched are not associated with the answer. Although GPT-4o ignores irrelevant information in SE ReDial prompts, it cannot do so in AAVE, showing its reasoning ability's brittleness.

\textbf{Failure to compose the program and execute all steps.} GPT-4o sometimes simulates an algorithm to solve math problems. However, it gets stuck in one step of its reasoning chain where it confuses the reference of comparison. For instance, in a question 'On Saturday, he sold … 4 fewer boxes of apple pie, than on Sunday. Come Sunday he done sold 5 more boxes of gingerbread than on Saturday and 15 more boxes of apple pie.' GPT-4o reasons by starting with 'Let $A_s$
 be the number of boxes of apple pie sold on Saturday...Let 
$A_u$ be the number of boxes of apple pie sold on Sunday'. Then, it realizes there is a contradiction between the correct formula $A_s = A_u - 4$
 and the wrong formula $A_u = A_s + 15$
, where it wrongly considers the number of apple pies sold on Sunday to be 15 more than that on Saturday. This indicates that reasoning with queries expressed in dialects limits a model’s reasoning ability. 

\subsection{Synthetic Lexical Feature list}
\label{sec:synthetic_lexical_feature}
Following \cite{ziems2022value}, we implement a feature list with distinct AAVE lexicons: ['got', 'ain't', 'no', "'", 'gonna', 'wanna', 'gotta', 'done', 'been', 'finna', 'gunna', 'gon',], and compute the Spearman's correlation between their frequencies in AAVE ReDial and the performance drop.

\subsection{Statement of Contribution}
All co-authors contributed to discussions, provided input on various aspects of the project, and assisted with writing, editing, and advising. In addition to these contributions, FL developed the initial idea, designed and conducted the experiments, contributed significantly to data collection, drafted the paper, and performed all analyses unless otherwise specified. As FL's mentors during her internship at Microsoft, SM and AW contributed significantly by coordinating resources, guiding the overall direction of the project including data collection, managing the ethical review process, and serving as the primary corresponding authors. XW conducted experiments on Claude and GPT-o1. On top of advising and paper writing, ELM developed some of the initial experiments for the ablation study and VH contributed to ideation and experiment design.

\clearpage
\newpage
\onecolumn
\subsection{Full Results on ReDial}
\label{sec:full_results}
We present the complete results on Redial. Specifically, Table~\ref{tab:all_res_algorithm_gold} provides the detailed results for Algorithm, Table~\ref{tab:all_res_logic} covers the results for Logic, Table~\ref{tab:all_res_math_gold} reports the results for Math, and Table~\ref{tab:all_res_comprehensive} reports the results for Integrated Tasks.
\begin{table*}[ht]
    \centering
    
    \begin{tabular}{c|c|c|c|c|c}
    \toprule
        Model&Setting&\multicolumn{2}{|c}{HumanEval}&\multicolumn{2}{|c}{MBPP}\\\cline{3-6}
        &&Original&AAVE&Original&AAVE\\
        \midrule
GPT-o1 \faLock&Vanilla&0.860 & 0.860$_{(+) 0.000}$ & 0.773 & 0.787$_{(+) 0.013}$ \\
\multirow{2}{*}{GPT-4o \faLock}&Vanilla&0.872&0.811$_{(-)0.061}$&0.700&0.707$_{(+)0.007}$\\
&CoT&0.841&0.805$_{(-)0.037}$&0.693&0.713$_{(+)0.02}$\\
\multirow{2}{*}{GPT-4 \faLock}&Vanilla&0.780&0.744$_{(-)0.037}$&0.700&0.700$_{(-)-0.0}$\\
&CoT&0.750&0.707$_{(-)0.043}$&0.693&0.500$_{(-)0.193}$\\
\multirow{2}{*}{GPT-3.5-turbo \faLock}&Vanilla&0.640&0.622$_{(-)0.018}$&0.667&0.640$_{(-)0.027}$\\
&CoT&0.616&0.591$_{(-)0.024}$&0.680&0.507$_{(-)0.173}$\\\hline
\multirow{2}{*}{Claude-Sonnet \faLock}&Vanilla&0.787 & 0.848$_{(+) 0.061}$ & 0.753 & 0.760$_{(+) 0.007}$ \\
&CoT&0.793 & 0.726$_{(-) 0.067}$ & 0.753 & 0.747$_{(-) 0.007}$ \\\hline
\multirow{2}{*}{Llama-3.1-70B-Instruct}&Vanilla&0.744&0.726$_{(-)0.018}$&0.707&0.573$_{(-)0.133}$\\
&CoT&0.738&0.689$_{(-)0.049}$&0.707&0.613$_{(-)0.093}$\\
\multirow{2}{*}{Llama-3-70B-Instruct}&Vanilla&0.689&0.671$_{(-)0.018}$&0.673&0.613$_{(-)0.06}$\\
&CoT&0.720&0.665$_{(-)0.055}$&0.673&0.627$_{(-)0.047}$\\
\multirow{2}{*}{Llama-3-8B-Instruct}&Vanilla&0.530&0.524$_{(-)0.006}$&0.540&0.493$_{(-)0.047}$\\
&CoT&0.537&0.512$_{(-)0.024}$&0.527&0.440$_{(-)0.087}$\\
\hline
\multirow{2}{*}{Mixtral-8x7B-Instruct-v0.1}&Vanilla&0.402&0.390$_{(-)0.012}$&0.507&0.413$_{(-)0.093}$\\
&CoT&0.396&0.396$_{(-)-0.0}$&0.547&0.427$_{(-)0.12}$\\
\multirow{2}{*}{Mistral-7B-Instruct-v0.3}&Vanilla&0.268&0.268$_{(-)-0.0}$&0.400&0.240$_{(-)0.16}$\\
&CoT&0.262&0.274$_{(+)0.012}$&0.367&0.213$_{(-)0.153}$\\
\hline
\multirow{2}{*}{Phi-3-Medium-128K-Instruct}&Vanilla&0.530&0.518$_{(-)0.012}$&0.560&0.340$_{(-)0.22}$\\
&CoT&0.530&0.573$_{(+)0.043}$&0.567&0.327$_{(-)0.24}$\\
\multirow{2}{*}{Phi-3-Small-128K-Instruct}&Vanilla&0.598&0.329$_{(-)0.268}$&0.633&0.167$_{(-)0.467}$\\
&CoT&0.585&0.293$_{(-)0.293}$&0.553&0.087$_{(-)0.467}$\\
\multirow{2}{*}{Phi-3-Mini-128K-Instruct}&Vanilla&0.549&0.482$_{(-)0.067}$&0.567&0.367$_{(-)0.2}$\\
&CoT&0.567&0.530$_{(-)0.037}$&0.587&0.347$_{(-)0.24}$\\
\hline
    \bottomrule
        
    \end{tabular}
    \caption{All results for \textbf{Algorithm}.}
    \label{tab:all_res_algorithm_gold}
\end{table*}

\begin{table}[ht]
    \centering

    \begin{tabular}{c|c|c|c|c|c}
    \toprule
        Model&Setting&\multicolumn{2}{|c}{Folio}&\multicolumn{2}{|c}{LogicBench}\\\cline{3-6}
        &&Original&AAVE&Original&AAVE\\
        \midrule
        GPT-o1 \faLock&Vanilla&0.963 & 0.938$_{(-) 0.025}$ & 0.810 & 0.715$_{(-) 0.095}$ \\
         \multirow{2}{*}{GPT-4o \faLock}&Vanilla&0.938&0.870$_{(-)0.068}$&0.720&0.685$_{(-)0.035}$\\
&CoT&0.938&0.926$_{(-)0.012}$&0.715&0.645$_{(-)0.070}$\\
\multirow{2}{*}{GPT-4 \faLock}&Vanilla&0.858&0.796$_{(-)0.062}$&0.745&0.710$_{(-)0.035}$\\
&CoT&0.864&0.759$_{(-)0.105}$&0.735&0.730$_{(-)0.005}$\\
\multirow{2}{*}{GPT-3.5-turbo \faLock}&Vanilla&0.605&0.519$_{(-)0.086}$&0.475&0.565$_{(+)0.090}$\\
&CoT&0.519&0.506$_{(-)0.012}$&0.490&0.360$_{(-)0.130}$\\
\hline
\multirow{2}{*}{Claude-Sonnet \faLock}&Vanilla&0.914 & 0.895$_{(-) 0.019}$ & 0.800 & 0.680$_{(-) 0.120}$ \\
&CoT&0.907 & 0.877$_{(-) 0.031}$ & 0.820 & 0.730$_{(-) 0.090}$ \\
\hline
\multirow{2}{*}{Llama-3.1-70B-Instruct}&Vanilla&0.642&0.593$_{(-)0.049}$&0.750&0.660$_{(-)0.090}$\\
&CoT&0.870&0.827$_{(-)0.043}$&0.760&0.720$_{(-)0.040}$\\
\multirow{2}{*}{Llama-3-70B-Instruct}&Vanilla&0.673&0.623$_{(-)0.049}$&0.655&0.495$_{(-)0.160}$\\
&CoT&0.883&0.809$_{(-)0.074}$&0.400&0.360$_{(-)0.040}$\\
\multirow{2}{*}{Llama-3-8B-Instruct}&Vanilla&0.667&0.617$_{(-)0.049}$&0.325&0.340$_{(+)0.015}$\\
&CoT&0.599&0.660$_{(+)0.062}$&0.375&0.355$_{(-)0.020}$\\
\hline
\multirow{2}{*}{Mixtral-8x7B-Instruct-v0.1}&Vanilla&0.327&0.401$_{(+)0.074}$&0.485&0.110$_{(-)0.375}$\\
&CoT&0.370&0.284$_{(-)0.086}$&0.395&0.285$_{(-)0.110}$\\
\multirow{2}{*}{Mistral-7B-Instruct-v0.3}&Vanilla&0.481&0.537$_{(+)0.056}$&0.180&0.055$_{(-)0.125}$\\
&CoT&0.475&0.506$_{(+)0.031}$&0.200&0.120$_{(-)0.080}$\\
\hline
\multirow{2}{*}{Phi-3-Medium-128K-Instruct}&Vanilla&0.543&0.568$_{(+)0.025}$&0.465&0.390$_{(-)0.075}$\\
&CoT&0.698&0.574$_{(-)0.123}$&0.325&0.330$_{(+)0.005}$\\
\multirow{2}{*}{Phi-3-Small-128K-Instruct}&Vanilla&0.580&0.531$_{(-)0.049}$&0.490&0.520$_{(+)0.030}$\\
&CoT&0.728&0.568$_{(-)0.160}$&0.395&0.485$_{(+)0.090}$\\
\multirow{2}{*}{Phi-3-Mini-128K-Instruct}&Vanilla&0.420&0.352$_{(-)0.068}$&0.755&0.665$_{(-)0.090}$\\
&CoT&0.481&0.370$_{(-)0.111}$&0.735&0.655$_{(-)0.080}$\\
\hline

    \end{tabular}
    \caption{All results for \textbf{Logic}.}
    \label{tab:all_res_logic}
\end{table}

\begin{table}[ht]
    \centering

    \begin{tabular}{c|c|c|c|c|c}
    \toprule
        Model&Setting&\multicolumn{2}{|c}{GSM8K}&\multicolumn{2}{|c}{SVAMP}\\\cline{3-6}
        &&Original&AAVE&Original&AAVE\\
        \midrule\hline
GPT-o1 \faLock&Vanilla&0.953 & 0.927$_{(-) 0.027}$ & 0.940 & 0.920$_{(-) 0.020}$ \\
\multirow{2}{*}{GPT-4o \faLock}&Vanilla&0.933&0.947$_{(+)0.013}$&0.933&0.913$_{(-)0.020}$\\
&CoT&0.967&0.933$_{(-)0.033}$&0.933&0.907$_{(-)0.027}$\\
\multirow{2}{*}{GPT-4 \faLock}&Vanilla&0.840&0.640$_{(-)0.200}$&0.840&0.787$_{(-)0.053}$\\
&CoT&0.947&0.867$_{(-)0.080}$&0.893&0.760$_{(-)0.133}$\\
\multirow{2}{*}{GPT-3.5-turbo \faLock}&Vanilla&0.587&0.287$_{(-)0.300}$&0.747&0.600$_{(-)0.147}$\\
&CoT&0.780&0.480$_{(-)0.300}$&0.727&0.607$_{(-)0.120}$\\
\hline
\multirow{2}{*}{Claude-Sonnet \faLock}&Vanilla&0.973 & 0.947$_{(-) 0.027}$ & 0.967 & 0.913$_{(-) 0.053}$ \\
&CoT&0.973 & 0.960$_{(-) 0.013}$ & 0.933 & 0.920$_{(-) 0.013}$ \\
\hline
\multirow{2}{*}{Llama-3.1-70B-Instruct}&Vanilla&0.680&0.920$_{(+)0.240}$&0.853&0.867$_{(+)0.013}$\\
&CoT&0.867&0.927$_{(+)0.060}$&0.893&0.813$_{(-)0.080}$\\
\multirow{2}{*}{Llama-3-70B-Instruct}&Vanilla&0.933&0.920$_{(-)0.013}$&0.880&0.853$_{(-)0.027}$\\
&CoT&0.947&0.907$_{(-)0.040}$&0.900&0.867$_{(-)0.033}$\\
\multirow{2}{*}{Llama-3-8B-Instruct}&Vanilla&0.847&0.800$_{(-)0.047}$&0.807&0.800$_{(-)0.007}$\\
&CoT&0.820&0.800$_{(-)0.020}$&0.833&0.800$_{(-)0.033}$\\
\hline
\multirow{2}{*}{Mixtral-8x7B-Instruct-v0.1}&Vanilla&0.427&0.193$_{(-)0.233}$&0.613&0.487$_{(-)0.127}$\\
&CoT&0.673&0.573$_{(-)0.100}$&0.700&0.560$_{(-)0.140}$\\
\multirow{2}{*}{Mistral-7B-Instruct-v0.3}&Vanilla&0.367&0.147$_{(-)0.220}$&0.433&0.280$_{(-)0.153}$\\
&CoT&0.420&0.320$_{(-)0.100}$&0.487&0.373$_{(-)0.113}$\\
\hline
\multirow{2}{*}{Phi-3-Medium-128K-Instruct}&Vanilla&0.893&0.833$_{(-)0.060}$&0.840&0.747$_{(-)0.093}$\\
&CoT&0.893&0.853$_{(-)0.040}$&0.827&0.800$_{(-)0.027}$\\
\multirow{2}{*}{Phi-3-Small-128K-Instruct}&Vanilla&0.840&0.793$_{(-)0.047}$&0.800&0.727$_{(-)0.073}$\\
&CoT&0.880&0.873$_{(-)0.007}$&0.907&0.813$_{(-)0.093}$\\
\multirow{2}{*}{Phi-3-Mini-128K-Instruct}&Vanilla&0.520&0.573$_{(+)0.053}$&0.520&0.527$_{(+)0.007}$\\
&CoT&0.800&0.807$_{(+)0.007}$&0.747&0.693$_{(-)0.053}$\\
\hline
    \bottomrule
    \end{tabular}
    \caption{All results for \textbf{Math}.}
    \label{tab:all_res_math_gold}
\end{table}

\begin{table}[ht]
    \centering
    \begin{tabular}{c|c|c|c}
    \toprule
         Model&Setting& Original &AAVE \\\hline\midrule
          GPT-o1 \faLock&Vanilla&0.942 & 0.925$_{(-) 0.017}$ \\
         \multirow{2}{*}{GPT-4o \faLock}&Vanilla&0.783&0.312$_{(-)0.471}$\\
&CoT&0.762&0.662$_{(-)0.1}$\\
\multirow{2}{*}{GPT-4 \faLock}&Vanilla&0.217&0.133$_{(-)0.083}$\\
&CoT&0.283&0.058$_{(-)0.225}$\\
\multirow{2}{*}{GPT-3.5-turbo \faLock}&Vanilla&0.200&0.129$_{(-)0.071}$\\
&CoT&0.075&0.067$_{(-)0.008}$\\
\hline
\multirow{2}{*}{Claude-Sonnet \faLock}&Vanilla&0.879 & 0.717$_{(-) 0.162}$ \\
&CoT&0.900 & 0.771$_{(-) 0.129}$ \\
\hline
\multirow{2}{*}{Llama-3.1-70B-Instruct}&Vanilla&0.392&0.113$_{(-)0.279}$\\
&CoT&0.579&0.500$_{(-)0.079}$\\
\multirow{2}{*}{Llama-3-70B-Instruct}&Vanilla&0.158&0.067$_{(-)0.092}$\\
&CoT&0.517&0.350$_{(-)0.167}$\\
\multirow{2}{*}{Llama-3-8B-Instruct}&Vanilla&0.025&0.067$_{(+)0.042}$\\
&CoT&0.029&0.025$_{(-)0.004}$\\
\hline
\multirow{2}{*}{Mixtral-8x7B-Instruct-v0.1}&Vanilla&0.100&0.075$_{(-)0.025}$\\
&CoT&0.133&0.071$_{(-)0.062}$\\
\multirow{2}{*}{Mistral-7B-Instruct-v0.3}&Vanilla&0.096&0.075$_{(-)0.021}$\\
&CoT&0.083&0.083$_{(-)-0.0}$\\
\hline
\multirow{2}{*}{Phi-3-Medium-128K-Instruct}&Vanilla&0.050&0.037$_{(-)0.013}$\\
&CoT&0.067&0.029$_{(-)0.037}$\\
\multirow{2}{*}{Phi-3-Small-128K-Instruct}&Vanilla&0.058&0.062$_{(+)0.004}$\\
&CoT&0.096&0.079$_{(-)0.017}$\\
\multirow{2}{*}{Phi-3-Mini-128K-Instruct}&Vanilla&0.021&0.042$_{(+)0.021}$\\
&CoT&0.017&0.021$_{(+)0.004}$\\
\hline
    \end{tabular}
    \caption{All results for \textbf{Integrated}.}
    \label{tab:all_res_comprehensive}
\end{table}

\clearpage
\newpage
\twocolumn

\end{document}